\documentclass[10pt,twocolumn,letterpaper]{article}

\usepackage[pagenumbers]{cvpr} 

\usepackage[dvipsnames,table]{xcolor}

\usepackage{algorithmic}
\usepackage{amsmath}
\usepackage{amssymb}
\usepackage{cite}
\usepackage{booktabs}
\usepackage{bm}
\usepackage{cite}
\usepackage{amsfonts}
\usepackage{textcomp}
\usepackage{makecell}
\usepackage{multirow}
\definecolor{dino}{RGB}{249,231,227}

\usepackage{setspace}
\usepackage[ruled,norelsize]{algorithm2e}

\definecolor{commentcolor}{RGB}{110,154,155}   
\newcommand{\PyComment}[1]{\ttfamily\textcolor{commentcolor}{\# #1}}  
\newcommand{\PyCode}[1]{\ttfamily\textcolor{black}{#1}} 

\definecolor{cvprblue}{rgb}{0.21,0.49,0.74}
\usepackage[pagebackref,breaklinks,colorlinks,citecolor=cvprblue]{hyperref}

\def\paperID{4372} 
\def\confName{CVPR}
\def\confYear{2024}

\title{
Feature Re-Embedding: Towards Foundation Model-Level Performance in Computational Pathology
}

\author{Wenhao Tang$^1$\footnotemark[2] ~~ Fengtao Zhou$^2$\footnotemark[2] ~~ Sheng Huang$^1$\footnotemark[1]  ~~ Xiang Zhu$^1$ ~~ Yi Zhang$^{1,4}$ ~~ Bo Liu$^3$\\
{ $^1$ Chongqing University ~~ $^2$ The Hong Kong University of Science and Technology} \\ {$^3$ Walmart Global Tech} {$^4$ Chongqing Normal University}\\
{\tt\small \{whtang, huangsheng, zhangyii\}@cqu.edu.cn, zhuxiang@stu.cqu.edu.cn} 
\\ 
{\tt\small fzhouaf@connect.ust.hk, kfliubo@gmail.com}
}

\begin{document}
\maketitle
\renewcommand{\thefootnote}{\fnsymbol{footnote}}
\footnotetext[1]{Corresponding Author.}
\footnotetext[2]{Equal Contribution.}
\begin{abstract}

Multiple instance learning (MIL) is the most widely used framework in computational pathology, encompassing sub-typing, diagnosis, prognosis, and more. However, the existing MIL paradigm typically requires an offline instance feature extractor, 
such as a pre-trained ResNet or a foundation model.
This approach lacks the capability for feature fine-tuning within the specific downstream tasks, limiting its adaptability and performance. To address this issue, we propose a Re-embedded Regional Transformer (R$^2$T) for re-embedding the instance features online, which captures fine-grained local features and establishes connections across different regions.
Unlike existing works that focus on pre-training powerful feature extractor or designing sophisticated instance aggregator,
R$^2$T is tailored to re-embed instance features online.
It serves as a portable module that can seamlessly integrate into mainstream MIL models. Extensive experimental results on common computational pathology tasks validate that:
1) feature re-embedding improves the performance of MIL models based on ResNet-50 features to the level of foundation model features, and further enhances the performance of foundation model features;
2) the R$^2$T can introduce more significant performance improvements to various MIL models; 
3) R$^2$T-MIL, as an R$^2$T-enhanced AB-MIL, outperforms other latest methods by a large margin.
The code is available at:~\href{https://github.com/DearCaat/RRT-MIL}{https://github.com/DearCaat/RRT-MIL}.




\end{abstract}    
\section{Introduction}
\label{sec:intro}
\begin{figure}[t]
    \begin{center}
    \includegraphics[width=8.5cm]{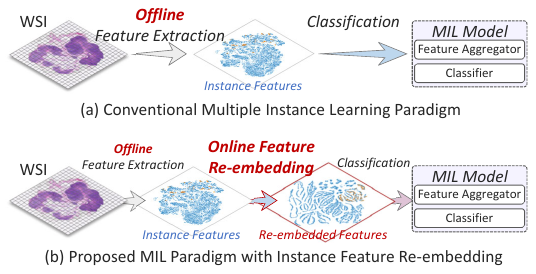}
    \end{center}
    \vspace{-0.3cm}
    \caption{\textbf{Top:} The conventional MIL paradigm lacks fine-tuning of the offline embedded instance features. \textbf{Bottom:} The proposed MIL paradigm that introduces instance feature re-embedding to provide more discriminative features for the MIL model.}
    \vspace{-0.3cm}
    \label{fig:intro}
\end{figure}

Computational pathology~\cite{cui2021artificial, song2023artificial, cifci2023ai} is an interdisciplinary field that combines pathology, image analysis, and computer science to develop and apply computational methods for the analysis and interpretation of pathological images (also known as whole slide images, WSIs). This field utilizes advanced algorithms, machine learning, and artificial intelligence techniques to assist pathologists in tasks like sub-typing~\cite{ilse2018attention}, diagnosis~\cite{zhang2022dtfd,lin2023interventional}, prognosis~\cite{wen2023deep,yao2020whole}, and more. However, the process of pixel-level labeling in ultra-high resolution WSIs is time-consuming and labor-intensive, posing challenges for traditional deep learning methods that rely on pixel-level labels in computational pathology. 
To address this challenge, multiple instance learning (MIL) approaches have been employed to treat WSI analysis as a weakly supervised learning problem~\cite{wang2019weakly, lu2021data}. MIL divides each WSI (referred to as a bag) into numerous image patches or instances. Previous MIL-based methods mainly follow a three-step process: 1) instance feature extraction, 2) instance feature aggregation, and 3) bag prediction. However, most previous works focus on the last two steps, where the extracted offline instance features are utilized to make bag-level predictions. 

Despite achieving ``clinical-grade" performance on numerous computational pathology tasks~\cite{chen2022hipt, zhang2022dtfd}, the conventional MIL paradigm faces a significant design challenge due to the large number of instances involved. The holistic end-to-end learning of the instance feature extractor, instance-level feature aggregator, and bag-level predictor becomes infeasible due to the prohibitively high memory cost. In previous works, an offline feature extractor pre-trained on natural images is used to extract instance features. However, this approach lacks a feature fine-tuning process for specific downstream tasks~\cite{clam, shao2021transmil, zhang2022dtfd}, resulting in low discriminative features and sub-optimal performance, as illustrated in Figure~\ref{fig:intro}(a). To mitigate this issue, some works~\cite{li2021dual, chen2022hipt, huang2023plip} have employed self-supervised methods to pre-train a more powerful feature extractor on a massive amount of WSIs, which are known as foundation models. Nevertheless, pre-training foundation model requires huge amounts of data ($>$200k WSIs) and computational resources. \textbf{Furthermore, the challenge of lacking feature fine-tuning remains unresolved.} An intuitive way of addressing the issue is to perform online features re-embedding using representation learning techniques before MIL models. As shown in Figure~\ref{fig:intro}(b), 
re-embedding modules can be trained end-to-end with MIL models to provide supervised feature fine-tuning. It enables fully exploiting the knowledge beneficial to the final task.

As a powerful representation learning method, Transformer~\cite{vaswani2017attention} has proven to be effective for representation learning and has demonstrated promising results in various domains~\cite{chen2022structure, zerveas2021transformer, yang2021graphformers}. However, directly applying the existing Transformers for re-embedding is challenging due to the characteristics of WSI.
The main problem is the unacceptable memory consumption caused by the massive input of image patches.
The linear multi-head self-attention (MSA)~\cite{xiong2021nystromformer} can alleviate the memory dilemma, but suffers from high computational cost and sub-optimal performance.
Moreover, the global MSA fails to capture the local detail features that are crucial for computational pathology.

In this paper, we propose Re-embedded Regional Transformer (R$^2$T), which leverages the advantages of the native MSA while overcoming its limitations. Specifically, R$^2$T applies the native MSA to each local region separately. Then, it uses a Cross-region MSA (CR-MSA) to fuse the information from different regions. Finally, a novel Embedded Position Encoding Generator (EPEG) is used to effectively encode the positional information of the patches.
By incorporating with mainstream MIL models, the proposed R$^2$T can ensure efficient computation while 
maintaining powerful representation capabilities to 
fine-tune the offline features according to the specific downstream tasks.
The main contributions can be summarized as follows:
\begin{itemize}
    \item We propose a novel paradigm for MIL models that incorporates a re-embedding module to address the issue of poor discriminative ability in instance features caused by offline feature extractors. The proposed feature re-embedding fashion can effectively improve MIL models, even achieving competitive performance compared to the latest foundation model.
    \item For re-embedding instance features, we design a Re-embedded Regional Transformer (R$^2$T) which can be seamlessly integrated into mainstream MIL models to further improve performance. By incorporating the R$^2$T into AB-MIL, we present the R$^2$T-MIL, which achieves state-of-the-art performance on various computational pathology benchmarks.
    \item We introduce two novel components for the R$^2$T: the  Cross-region MSA and the Embedded Position Encoding Generator. The former enables effective information fusion across different regions. The latter combines the benefits of relative and convolutional position encodings to encode the positional information more effectively.
\end{itemize}

\section{Related Work}
\subsection{Computational Pathology}
The transition from traditional glass slides to digital pathology has provided a wealth of opportunities for computational pathology, which aims to combine pathology, image analysis, and computer science techniques to develop computer-assisted methods for analyzing pathology images~\cite{cui2021artificial, song2023artificial, cifci2023ai}. By harnessing the power of advanced machine learning algorithms, computational pathology can enable large-scale data analysis and facilitate collaboration among pathologists and researchers. Traditionally, pathologists relied on visual examination of tissue samples under a microscope to make diagnoses. However, this manual process was not only time-consuming but also prone to subjective interpretations and human errors. With the emergence of computational pathology, these limitations are being addressed in remarkable ways. By automating labor-intensive processes, it can liberate pathologists' time, enabling them to focus on complex and critical decision-making tasks. Meanwhile, its ability to leverage vast amounts of data, combined with advanced analytic, holds great promise for breakthroughs in personalized medicine. By extracting quantitative features from pathology images, computational pathology can assist in making diagnosis~\cite{li2021dual, clam, ilse2018attention}, predicting patient outcomes~\cite{zhu2017wsisa, wulczyn2020deep}, identifying biomarkers~\cite{xie2021evaluating, hamilton2019digital}, and guiding tailored treatment strategies~\cite{tran2019personalized}. 

\subsection{Multiple Instance Learning}
Multiple instance learning (MIL) is the most widely used paradigm in computational pathology, involving three key steps: slide patching, instance feature extraction, and bag label prediction~\cite{ilse2018attention, campanella2019clinical, clam}. Due to the ultra-high resolution of WSIs, the instance features are typically extracted by pre-trained models, especially ResNet-50 pre-trained on ImageNet. However, the inherent difference between pathology images and nature images results in poor discrimination of extracted features. Some self-supervised learning-based methods~\cite{chen2022hipt, li2021dual, zhao2020predicting, saillard2021self,huang2023plip} attempt to alleviate the feature bias by pre-training feature extractor on a large number of WSIs. 
For example, Huang~et al. adapted CLIP~\cite{clip} to pre-train a vision Transformer called PLIP, with more than 200k slide-text pairs~\cite{huang2023plip}.
These efforts aim to enhance the discrimination of offline features by leveraging the vast amount of pathology-specific information available in the pre-training data. 
The extracted instance features are then utilized for bag prediction in computational pathology. These methods can be categorized into instance label fusion~\cite{xu2019camel, campanella2019clinical, instance_mil_1, instance_mil_2} and instance feature fusion~\cite{li2021dual, clam, shao2021transmil, zhang2022dtfd, sharma2021cluster}. Instance label fusion methods first obtain instance labels and then pool them to obtain the bag label, while instance feature fusion methods aggregate all instance features into a high-level bag embedding and then obtain the bag prediction. Recently, Transformer blocks~\cite{vaswani2017attention} have been utilized to aggregate instance features~\cite{shao2021transmil, li2021dt, wang2022lymph}, demonstrating the advantage of self-attention over traditional attention~\cite{li2021dual, clam, ilse2018attention} in modeling mutual instance information. 
While existing methods in computational pathology have shown promising results, most of them primarily focus on how to aggregate discriminative information from pre-extracted features. However, the pre-extracted features lack fine-tuning on specific downstream tasks, resulting in sub-optimal performance.

\begin{figure*}
\centerline{\includegraphics[width=0.9\textwidth]{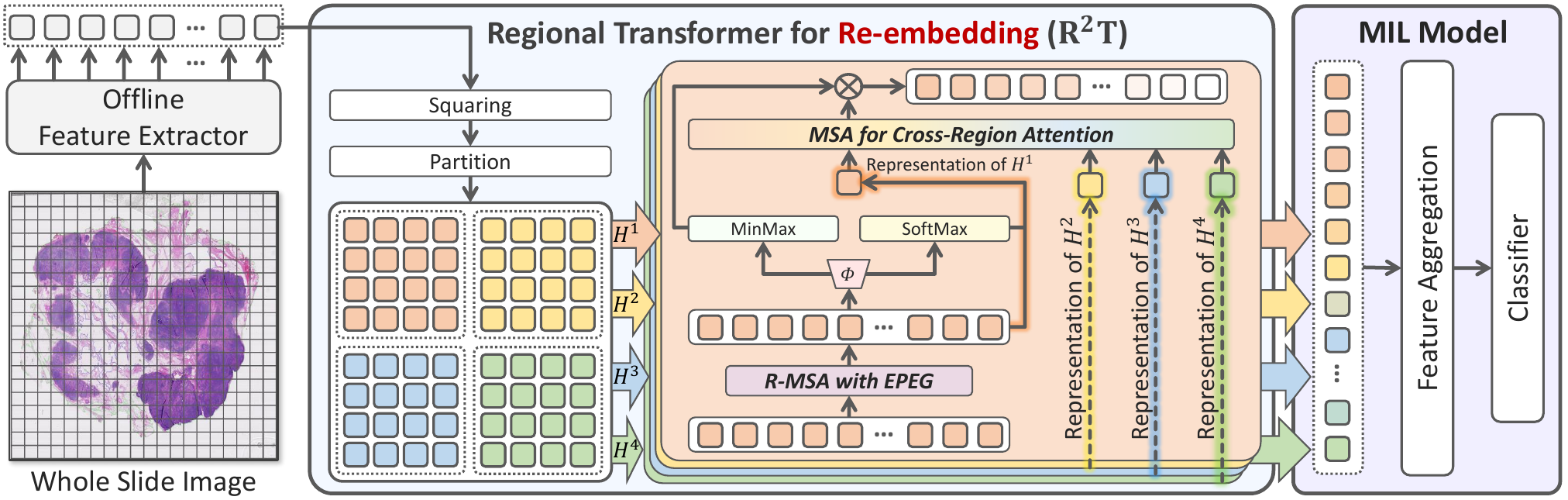}}
\caption{Overview of proposed R$^2$T-MIL. A set of patches is first cropped from the tissue regions of a slide and embedded in features by an offline extractor. Then, the sequence is processed with the R$^2$T module: (1) region partition, (2) feature re-embedding within each region, and (3) cross-region feature fusion. Finally, a MIL model predicts the bag labels using the re-embedded instance features.}
\label{fig:model}
\end{figure*}
\section{Methodology}
\subsection{Preliminary}
From the perspective of MIL, a WSI $X$ is considered as a bag while its patches are deemed as instances in this bag, which can be represented as $X=\{x_i\}^I_{i=1}$. The instance number $I$ varies for different bags. For a classification task, there exists a known label $Y$ for a bag and an unknown label $y_i$ for each of its instances. If there is at least one positive instance in a bag, then this bag is positive; otherwise, it is negative.
The goal of a MIL model $\mathcal{M}(\cdot)$ is to predict the bag label with all instances $\hat{Y}\leftarrow\mathcal{M}(X)$.
Following the recent popular approaches~\cite{xu2019camel,instance_mil_2}, the MIL prediction process can be divided into three steps: instance feature extraction, instance feature aggregation, and bag classification. Specifically, this process can be defined as follows:
\begin{equation}
\hat{Y}\leftarrow \mathcal{M}(X):=\mathcal{C}(\mathcal{A}(\mathcal{F}(X))),
\end{equation}
where $\mathcal{F}(\cdot)$, $\mathcal{A}(\cdot)$, and $\mathcal{C}(\cdot)$ are the mapping functions of these aforementioned steps respectively.

In computational pathology, extracting all instances in a bag poses a huge computational challenge for the end-to-end optimization of these three steps.
Therefore, most existing approaches rely on a pre-trained deep learning model to obtain instance features first. 
Then, they only optimize the aggregation and classification steps. 
\textbf{However, the non-fine-tuned features lead to sub-optimal performance, even if the features are extracted by a foundation model.}
An intuitive way to address this problem is by re-embedding based on the extracted instance features, 
while most of the existing approaches pay more attention to feature aggregation and neglect the importance of re-embedding. 
In this paper, we 
include a re-embedding step after the instance feature extraction and update the bag labeling as follows,
\begin{equation}
\hat{Y}\leftarrow \mathcal{M}(X):=\mathcal{C}(\mathcal{A}(\mathcal{R}(\mathcal{F}(X)))),
\end{equation}
where $\mathcal{R}(\cdot)$ is the mapping function of the re-embedding.

\subsection{Re-embedded Regional Transformer}
As illustrated in Figure~\ref{fig:model}, we propose a Re-embedded Regional Transformer (R$^2$Transformer, R$^2$T) to re-embed the input instance features as the new instance representations,
\begin{equation}
    Z= \{z_i\}_{i=1}^I:=\mathcal{R}(H)\in \mathbb{R}^{I\times D},
\end{equation}
where $\mathcal{R}(\cdot)$ is the mapping function of the R$^2$Transformer here, $H=\{h_i\}_{i=1}^I:=\mathcal{F}(X)\in \mathbb{R}^{I\times D}$ is the processed input instance features, and $z_i=\mathcal{R}(h_i)$ is the new embedding of the $i$-th instance. $D$ is the dimension of the embedding. The R$^2$Transformer can be flexibly plugged into the MIL framework as a re-embedding module after feature input and before instance aggregation to reduce bias caused by the shift between offline feature learning and downstream tasks. The whole re-embedding process in the R$^2$Transformer can be formulated as,
\begin{equation}
\begin{aligned}
& \hat{Z}= \textrm{R-MSA}\left ( \textrm{LN}\left ( H \right ) \right ) + H \\
& Z = \textrm{CR-MSA}\left ( \textrm{LN}\left ( \hat{Z} \right ) \right ) + \hat{Z}
\end{aligned}
\end{equation}
where R-MSA$(\cdot)$ denotes Regional Multi-head Self-attention, CR-MSA$(\cdot)$ denotes Cross-region MSA, and LN$(\cdot)$ denotes Layer Normalization.

\noindent\textbf{Regional Multi-head Self-attention:}
Since instance number $I$ is very large, most Transformers in this field commonly adopt two strategies to avoid the Out-of-Memory issue. The first method is to sample or aggregate the original large instance set into a small one, after which global self-attention is performed~\cite{li2021dt,zhao2022setmil}. The second method performs the Nystrom algorithm~\cite{xiong2021nystromformer} to approximate the global self-attention~\cite{shao2021transmil}. Although these methods address the scalability issue of self-attention with a large $I$, they neglect the fact that the tumor areas are local and only occupy a small part of the whole image. Performing global self-attention on all instances results in feature homogenization. Moreover, different from the conventional MIL application scenarios, the instances in each bag have ordinal relations in computational pathology due to the fact that they are all collected from the same slide in order. These facts motivate us to design the Regional Multi-head Self-attention (R-MSA) that divides the bag into several different regions and performs self-attention in each region separately. R-MSA takes into account the aforementioned WSI properties and makes use of instance ordinal relation information to reduce computation complexity and highlight salient local features.

In R-MSA, the input instance features are reshaped into a 2-D feature map, $ H \in \mathbb{R}^{I \times D} \rightarrow H \in \mathbb{R}^{\left \lceil \sqrt{I}  \right \rceil\times \left \lceil \sqrt{I}  \right \rceil \times D} $. And $L \times L$ regions are then divided evenly across the map in a non-overlapping manner, with each containing $M \times M$ instances where $L\times M= \lceil\sqrt{I}\rceil$.
For example, the region partition starts from the top-left instance, and an $8 \times 8$ feature map is evenly partitioned into $2 \times 2$ regions of size $4 \times 4 \left ( L=2, M=4 \right )$. \textit{We fix the number of regions $L$ rather than the size of regions $M$ to obtain $L \times L$ regions with adaptive size.} By default, $L$ is set to $8$.
Self-attention is computed within each local region.
The whole process of R-MSA can be denoted as,

\begin{equation}
\begin{aligned}
&\bm{Step~1:}~  H \in \mathbb{R}^{I \times D}\overset{\textrm{Squaring}}\longrightarrow H \in  \mathbb{R}^{L^2\times M^2 \times D}, \\
&\bm{Step~2:}~ H\overset{\textrm{Partition}}\longrightarrow \{H^l\}^{L^2}_{l=1}, H^l\in  \mathbb{R}^{M\times M \times D},\\
&\bm{Step~3:}~\hat{Z}:=\{\hat{Z}^l\}^{L^2}_{l=1},\hat{Z}^l=\mathcal{S}(H^l)\in  \mathbb{R}^{M\times M \times D},\\
\end{aligned}
\end{equation}
where $\mathcal{S}(\cdot)$ is vanilla multi-head self-attention with our proposed Embedded Position Encoding Generator (EPEG).

\begin{figure}[t]
    \centering
    \includegraphics[width=7cm]{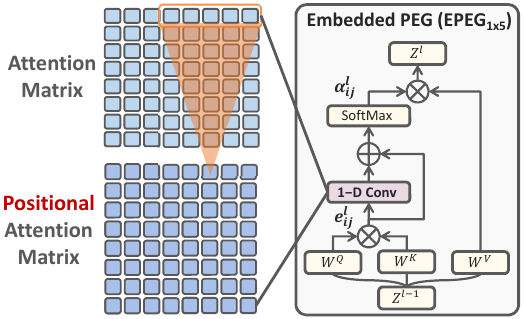}
    \caption{Illustration of Embedded Position Encoding Generator. }
    \vspace{-0.3cm}
    \label{fig:epeg}
\end{figure}

\begin{table*}[t]
\small
\centering
\begin{tabular}{llccccccc}
\toprule
\multicolumn{2}{c}{\multirow{2}{*}{Methods }}        & \multicolumn{3}{c}{CAMELYON-16}   && \multicolumn{3}{c}{TCGA-BRCA}  \\ \cline{3-5} \cline{7-9}
                                 && \makecell[c]{Accuracy}           & \makecell[c]{AUC}      & \makecell[c]{F1-score}
                                 && \makecell[c]{Accuracy}          & \makecell[c]{AUC}      & \makecell[c]{F1-score}   \\ \midrule

\multirow{8}{*}{\rotatebox{90}{\makecell[c]{ResNet-50\\ImageNet pretrained}}}&AB-MIL~\cite{ilse2018attention}  & 90.06$\pm$0.72             & 94.54$\pm$0.30           & 87.83$\pm$0.83
                                 && 86.41$\pm$4.92           & 91.10$\pm$2.52        & 81.64$\pm$4.71\\
&CLAM~\cite{clam}        & 90.14$\pm$0.85              & 94.70$\pm$0.76               & 88.10$\pm$0.63
                                 &&                85.17$\pm$2.70           & 91.67$\pm$1.78        & 80.37$\pm$3.04\\
&DSMIL~\cite{li2021dual}          & 90.17$\pm$1.02             & 94.57$\pm$0.40           & 87.65$\pm$1.18                                 && \underline{87.20$\pm$2.69}           & 91.58$\pm$1.33        & 82.41$\pm$2.92\\
&TransMIL~\cite{shao2021transmil} & 89.22$\pm$2.32              & 93.51$\pm$2.13   & 85.10$\pm$4.33
                                 && 84.68$\pm$2.67           & 90.80$\pm$1.91        & 79.86$\pm$2.63\\
&DTFD-MIL~\cite{zhang2022dtfd}    & 90.22$\pm$0.36              &  95.15$\pm$0.14           & 87.62$\pm$0.59
                                 && 85.92$\pm$1.76           & 91.43$\pm$1.64        & 81.09$\pm$2.05  \\
&IBMIL~\cite{lin2023interventional}    & 91.23$\pm$0.41              &  94.80$\pm$1.03           &88.80$\pm$0.89
                                 && 84.19$\pm$3.40           & 91.01$\pm$2.32        & 79.45$\pm$3.42\\
&MHIM-MIL~\cite{tang2023mhim}    & \underline{91.81$\pm$0.82}              &  \underline{96.14$\pm$0.52}           & \underline{89.94$\pm$0.70}
                                 && 86.73$\pm$5.59           & \underline{92.36$\pm$1.58}        & \underline{82.43$\pm$5.47}\\
& 
 \textbf{R$^2$T-MIL}        & \textbf{92.40$\pm$0.31}     & \textbf{97.32$\pm$0.29}      & \textbf{90.63$\pm$0.45}
                                && \textbf{88.33$\pm$0.67}           & \textbf{93.17$\pm$1.45}        & \textbf{83.70$\pm$0.95}\\
\midrule
\multirow{8}{*}{\rotatebox{90}{\makecell[c]{PLIP\\WSI pretrained}}}
&AB-MIL~\cite{ilse2018attention}  & 94.66$\pm$0.42& 97.30$\pm$0.31& 93.29$\pm$0.54&& 85.45$\pm$2.32           & 91.73$\pm$2.26        & 80.60$\pm$2.66         \\
&CLAM~\cite{clam}        & 93.73$\pm$0.54              & 97.17$\pm$0.50               & 91.60$\pm$0.60
                                 && 86.70$\pm$1.35           & 92.16$\pm$2.02        & 81.91$\pm$1.78   \\
&DSMIL~\cite{li2021dual}          & 94.40$\pm$0.85             & 97.06$\pm$0.56           & 92.78$\pm$1.15                                 && 87.25$\pm$2.70           & 91.80$\pm$1.67        & 82.18$\pm$2.28 \\
&TransMIL~\cite{shao2021transmil} & 94.40$\pm$0.43              & \underline{97.88$\pm$0.21}   & 92.81$\pm$0.43
                                 && 85.83$\pm$3.44           & 92.17$\pm$2.20        & 81.12$\pm$3.25  \\
&DTFD-MIL~\cite{zhang2022dtfd}    & 94.57$\pm$0.31              &  97.29$\pm$0.23           & 93.12$\pm$0.40
                                 && 86.42$\pm$2.67           & 92.16$\pm$2.42        & 81.77$\pm$2.73  \\
&IBMIL~\cite{lin2023interventional}    & 93.90$\pm$0.66 &  97.04$\pm$0.18           & {92.44$\pm$0.64}
                                 && \underline{87.57$\pm$1.48}           & 91.71$\pm$1.74        & \underline{82.78$\pm$2.02}\\
&MHIM-MIL~\cite{tang2023mhim}    & \underline{95.32$\pm$0.31}              &  97.79$\pm$0.15& \underline{94.13$\pm$0.42}
                                 && 87.07$\pm$2.20           & \underline{93.17$\pm$2.00}        & 82.48$\pm$2.50  \\
& 
 \textbf{R$^2$T-MIL}        & \textbf{95.49$\pm$0.00}     & \textbf{98.05$\pm$0.29}      & \textbf{94.29$\pm$0.04}
                                && \textbf{88.82$\pm$3.22}           & \textbf{93.80$\pm$1.24}        & \textbf{84.55$\pm$3.55} \\

\bottomrule
\end{tabular}
\caption{Cancer diagnosis, and sub-typing results on C16 and BRCA. The highest performance is in bold, and the second-best performance is underlined. 
With AB-MIL as a baseline, R$^2$T-MIL is not only capable of re-embedding ResNet-50 features to the level of foundation model (PLIP~\cite{huang2023plip}) features, but also effectively fine-tuning offline PLIP features.
}
\label{tab:main_comp}
\end{table*}

\noindent\textbf{Embedded Position Encoding Generator:}
Inspired by~\cite{shao2021transmil}, we adopt a convolutional computation called Position Encoding Generator (PEG)~\cite{chu2021peg} to address the challenge of traditional position encoding strategies being unable to handle input sequences of variable length.
Different from previous methods, we propose a novel approach called Embedded PEG (EPEG) by incorporating PEG into the R-MSA module, inspired by the relative position encoding strategy~\cite{wu2021irpe,liu2021swin,shaw2018rpe}. By embedding the PEG into the MSA module, EPEG can utilize a lightweight 1-D convolution $\textrm{Conv}_{\textrm{1-D}}(\cdot)$ to more effectively encode in each region separately.
The structure of EPEG is shown in Figure~\ref{fig:epeg}. Taking the instances in the $l$-th region as an example, the EPEG can be formulated as,
\begin{equation}
\alpha_{ij}^{l} = \mathrm{SoftMax} \left ( e_{ij}^{l} {\color{blue} + \textrm{Conv}_{\textrm{1-D}}\left ( e_{ij}^{l} \right ) } \right ),
\end{equation}
where $\alpha_{ij}^{l}$ is the attention weight of $H^{l}_{j}$ with respect to $H^{l}_{i}$, and $e_{ij}^{l}$ is calculated using a scaled dot-product attention.

\noindent\textbf{Cross-region Multi-head Self-attention:}
R-MSA only considers the features within each region, which limits its modeling power of context-based semantic features. This is crucial for downstream tasks, such as prognosis, which requires a more comprehensive judgment. To effectively model the cross-region connections, we propose Cross-region Multi-head Self-attention (CR-MSA). First, we aggregate the representative features $R^l$ of each region,
\begin{equation}
\centering
\begin{aligned}
 W^{l}_{a} &= \mathrm{SoftMax}_{m=1}^M\left ( \hat{Z}^{l}_{m}\Phi \right ), \\
 R^l &= W^{l\top}_a\hat{Z}^{l}, \\
\end{aligned}
\end{equation}
where $\Phi \in \mathbb{R}^{D\times K}$ denotes learnable parameters. We utilize vanilla MSA to model the cross-region connection, $\hat{R}= \mathcal{S}\left ( R \right )$. Finally, the updated representative features are distributed to each instance in the region with MinMax normalized weight $W^l_d$,
\begin{equation}
\centering
\begin{aligned}
W^l_d &= \mathrm{MinMax}_{m=1}^M\left ( \hat{Z}^l_m \Phi \right ) \in \mathbb{R}^{M^2\times K}, \\
Z^l &= W^{l\top}_d\hat{R}^{l}\hat{W}^{l}_d, \\
\end{aligned}
\end{equation}
where $\hat{W}^{l}_d = \mathrm{SoftMax}_{k=1}^K\left ( \hat{Z}^{l}_{mk}\Phi \right ) \in \mathbb{R}^{K\times 1}$.

\subsection{R$^2$Transformer-based MIL}
Once we obtain the re-embedding of instances, any instance aggregation method and classifier can be applied to accomplish the specific downstream tasks. The re-embedding $\mathcal{R}(\cdot)$ will be optimized with the instance aggregation module $\mathcal{A}(\cdot)$ and the bag classifier $\mathcal{C}(\cdot)$ together, 
\begin{equation}
    \{\mathcal{\hat{R}},\mathcal{\hat{A}},\mathcal{\hat{C}}\}\leftarrow \arg\min L(Y,\hat{Y})=L(Y,\mathcal{C}(\mathcal{A}(\mathcal{R}(H)))),
\end{equation}
where $L(\cdot,\cdot)$ denotes any MIL loss. 
R$^2$Transformer-based MIL (R$^2$T-MIL) adopts the instance aggregation method and the bag classifier of AB-MIL~\cite{ilse2018attention} by default.
\section{Experiments and Results}
\subsection{Datasets and Evaluation Metrics}
\noindent\textbf{Datasets:}
We use \textbf{CAMELYON-16~\cite{bejnordi2017diagnostic}} (C16), \textbf{TCGA-BRCA}, and \textbf{TCGA-NSCLC} to evaluate the performance on diagnosis and sub-typing tasks. For prognosis, we use \textbf{TCGA-LUAD}, \textbf{TCGA-LUSC}, \textbf{TCGA-BLCA} to evaluate the performance on the survival prediction task. Please refer to the Supplementary Material for more details.

\noindent\textbf{Evaluation Metrics:}
For diagnosis and sub-typing, we leverage Accuracy, Area Under Curve (AUC), and F1-score to evaluate model performance.
We only report AUC in ablation experiments.
For survival prediction, we report the C-index in all datasets.
To reduce the impact of data split on model evaluation, we follow~\cite{clam,Zhou_2023_ICCV,Zhang_2022_BMVC} and apply 5-fold cross-validation in all remaining datasets except C16.
We report the mean and standard deviation of the metrics over $N$ folds.
For C16, we follow~\cite{tang2023mhim} and use 3-times 3-fold cross-validation to alleviate the effects of random seed.

\noindent\textbf{Compared Methods:}
Seven influential MIL approaches are employed for comparison. They are AB-MIL~\cite{ilse2018attention}, DSMIL~\cite{li2021dual}, CLAM~\cite{clam}, DTFD-MIL~\cite{zhang2022dtfd}, TransMIL~\cite{shao2021transmil}, IBMIL~\cite{lin2023interventional}, and MHIM-MIL~\cite{tang2023mhim}. 
We reproduce the results of these methods under the same settings. 

\noindent\textbf{Implementation Details:}
We adopt ResNet50~\cite{he2016deep} pre-trained with ImageNet-1k and the latest foundation model PLIP~\cite{huang2023plip} pre-trained with OpenPath as the offline feature extractors.
Supplementary Material offers more details.

\subsection{Main Results}
\subsubsection{Cancer Diagnosis, and Sub-typing}

Table~\ref{tab:main_comp} presents the diagnosis and sub-typing performances of different MIL approaches on the C16 and BRCA datasets. The results demonstrate that our proposed R$^2$T-MIL achieves the best performance under all metrics on all benchmarks. Specifically, R$^2$T-MIL gets 0.59\%, 1.18\%, and 0.69\% performance gains over the second-best methods in Accuracy, AUC, and F1-score respectively on the C16 dataset. On the BRCA dataset, the AUC improvement is 0.69\%. R$^2$T-MIL employs the same aggregation and classification methods as AB-MIL. However, R$^2$T-MIL significantly outperforms AB-MIL. It increases the AUC by 2.78\% and 1.77\% on the C16 and BRCA datasets, respectively.
The sub-typing results on NSCLC in Table~\ref{tab:nsclc} support a similar observation. We attribute these substantial performance improvements to the additional re-embedding step based on our proposed R$^2$T, which surpasses the performance of the foundation model (+0.02\% AUC on C16, +1.37\% on BRCA, +0.72\% on NSCLC). In addition, we find that R$^2$T can further enhance the features of the foundation model, achieving considerable improvement. This validates the effectiveness of re-embedding. 

\begin{table}[tb]
\small
\centering
\begin{tabular}{p{0.15cm}lccc}
\toprule
\multicolumn{2}{c}{Methods }        & Accuracy & AUC & F1-score \\ \midrule
\multirow{8}{*}{\rotatebox{90}{ResNet-50}}&AB-MIL  & 90.32$_{\pm1.39}$& 95.29$_{\pm1.14}$& 89.83$_{\pm1.53}$\\
&CLAM        & 90.52$_{\pm2.08}$& 95.37$_{\pm1.08}$& 90.08$_{\pm1.97}$\\
&DSMIL          & 90.43$_{\pm2.52}$& 95.60$_{\pm0.81}$& 90.03$_{\pm2.61}$\\
&TransMIL & 90.04$_{\pm1.86}$& 94.97$_{\pm1.11}$& 89.94$_{\pm1.73}$\\
&DTFD-MIL    & 89.85$_{\pm1.53}$&  95.55$_{\pm1.47}$ & 89.60$_{\pm1.67}$\\
&IBMIL    & 90.04$_{\pm1.48}$ &  95.57$_{\pm1.13}$& 89.73$_{\pm1.64}$\\
&MHIM-MIL    & \underline{91.27$_{\pm2.35}$}              &  \underline{96.02$_{\pm1.35}$}           & \underline{90.85$_{\pm2.53}$}\\
& 
 \textbf{R$^2$T-MIL}        & \textbf{91.75$_{\pm2.38}$}     & \textbf{96.40$_{\pm1.13}$}      & \textbf{91.26$_{\pm2.60}$}\\
\midrule
\multirow{8}{*}{\rotatebox{90}{PLIP}}&AB-MIL  & 90.99$_{\pm2.43}$& 95.68$_{\pm1.98}$& 90.52$_{\pm2.45}$\\
&CLAM        & 90.80$_{\pm2.35}$& 95.46$_{\pm1.72}$& 90.38$_{\pm2.46}$\\
&DSMIL          & 90.60$_{\pm2.37}$& 95.78$_{\pm1.81}$& 90.24$_{\pm2.51}$\\
&TransMIL & 89.09$_{\pm2.00}$& 95.30$_{\pm1.95}$& 88.83$_{\pm2.16}$\\
&DTFD-MIL    & 90.42$_{\pm2.98}$&  95.83$_{\pm1.75}$ & 89.91$_{\pm3.01}$\\
&IBMIL    & 91.18$_{\pm3.27}$ &  95.62$_{\pm2.09}$& 90.94$_{\pm3.20}$\\
&MHIM-MIL    & \underline{91.74$_{\pm1.88}$}              &  \underline{96.21$_{\pm1.26}$}           & \underline{91.20$_{\pm1.89}$}\\
& 
 \textbf{R$^2$T-MIL}        & \textbf{92.13$_{\pm 2.55}$}     & \textbf{96.40$_{\pm 1.45}$}      & \textbf{91.83$_{\pm 2.50}$}\\
\bottomrule
\end{tabular}
\caption{Sub-typing results on TCGA-NSCLC. }
\label{tab:nsclc}
\end{table}

\begin{table}[tb]
\small
\centering
\begin{tabular}{p{0.15cm}lccc}
\toprule
\multicolumn{2}{c}{Methods }        & BLCA & LUAD & LUSC \\ \midrule
\multirow{8}{*}{\rotatebox{90}{ResNet-50}}&AB-MIL  & 57.50$_{\pm3.94}$             & 58.78$_{\pm4.90}$           & 56.51$_{\pm7.14}$ \\
&CLAM        & 57.57$_{\pm3.73}$              & 59.60$_{\pm3.93}$               & 56.65$_{\pm6.90}$\\
&DSMIL          & 57.42$_{\pm 2.25}$     & 59.31$_{\pm 4.75}$        & 55.03$_{\pm 6.61}$  \\
&TransMIL & \underline{58.90$_{\pm4.70}$}       & \underline{64.11$_{\pm1.99}$}      & 56.39$_{\pm2.94}$\\
&DTFD-MIL    & 56.98$_{\pm3.24}$      &  59.48$_{\pm2.61}$    & 55.16$_{\pm4.33}$\\
&IBMIL    & 58.41$_{\pm 2.90}$&  58.58$_{\pm 4.67}$           & \underline{59.18$_{\pm 3.29}$}\\
&MHIM-MIL    & 58.36$_{\pm3.26}$              &  60.32$_{\pm4.41}$           & 56.08$_{\pm6.33}$\\
& 
 \textbf{R$^2$T-MIL}        & \textbf{61.13$_{\pm2.36}$}     & \textbf{67.19$_{\pm 4.02}$}      & \textbf{60.95$_{\pm 4.41}$}\\
\midrule
\multirow{8}{*}{\rotatebox{90}{PLIP}}&AB-MIL  & 59.18$_{\pm2.48}$             & 62.09$_{\pm4.38}$           & 57.12$_{\pm2.39}$ \\
&CLAM        & \underline{61.58$_{\pm2.89}$}              & \underline{64.05$_{\pm4.70}$}               & 58.00$_{\pm3.34}$\\
&DSMIL          & 58.96$_{\pm 1.80}$     & 63.82$_{\pm 5.56}$       & 56.12$_{\pm 2.21}$ \\
&TransMIL & 56.20$_{\pm3.26}$       & 63.55$_{\pm2.94}$      & \underline{58.84$_{\pm3.28}$}\\
&DTFD-MIL    & 59.67$_{\pm4.71}$      & 61.78$_{\pm2.33}$   & 57.75$_{\pm3.52}$\\
&IBMIL    & 56.32$_{\pm 2.69}$              &  58.86$_{\pm 3.40}$           & 57.33$_{\pm 3.28}$\\
&MHIM-MIL    &     60.92$_{\pm 3.38}$           &  62.94$_{\pm4.58}$           & 55.95$_{\pm 2.54}$\\
& 
 \textbf{R$^2$T-MIL}        & \textbf{63.98$_{\pm2.26}$}     & \textbf{65.94$_{\pm1.34}$}      & \textbf{60.42$_{\pm2.15}$}\\
\bottomrule
\end{tabular}
\caption{Survival Prediction results on three main datasets. }
\label{tab:sp}
\end{table}
\subsubsection{Survival Prediction}

Table~\ref{tab:sp} shows the experimental results on three survival prediction datasets. It is worth noting that our proposed R$^2$T-MIL model demonstrates outstanding performance, attaining a C-index of 61.13\% on the BLCA, 67.19\% on the LUAD, and 60.95\% on the LUSC. It outperforms the compared methods by a significant margin, with improvements of 2.23\%, 3.08\%, and 1.77\% over the second-best methods, respectively. Furthermore, our proposed feature re-embedding strategy can yield substantial improvements even when working with high-quality features extracted by the foundation model. Particularly, compared to AB-MIL, our feature re-embedding strategy brings performance improvements of 4.8\%, 3.85\%, and 3.3\% for the three datasets, respectively. These results highlight the consistent and reliable performance of our proposed strategy and method, indicating its efficacy in predicting survival outcomes.

\begin{figure*}[t]
    \centering
    \includegraphics[width=16.5cm]{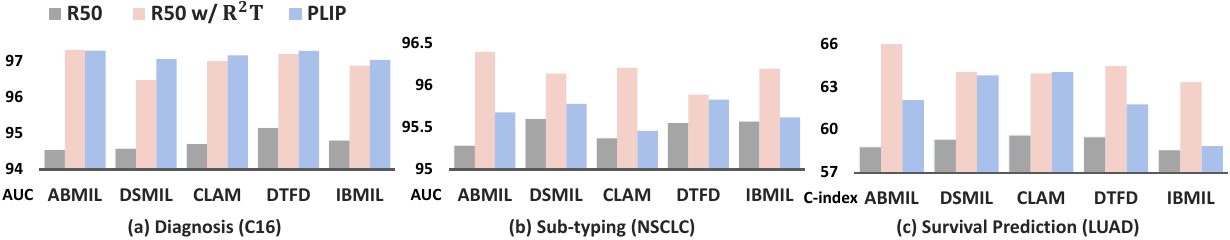}
    \caption{Performance improvement by adding R$^2$T.
    Features re-embedded by R$^2$T online outperform PLIP offline features on most tasks.
    }
    \label{fig:rt_bonus}
\end{figure*}

\subsection{Ablation Study}
\subsubsection{Re-Embedding Matters in Pathology}

\noindent\textbf{Foundation Model Features vs. Re-embedding Features:}
Table~\ref{tab:diff_trans} and Figure~\ref{fig:rt_bonus} compare the performance of the features extracted by the foundation model PLIP~\cite{huang2023plip} and various re-embedding features. The PLIP is based on the multi-modal foundation model CLIP~\cite{clip} and uses up to 200K slide-text pairs for pre-training. Although this high-cost pre-training brings some improvement on different tasks, it still has bottlenecks. We attribute this to the dilemma of the traditional paradigm that even the best offline pre-training features cannot address the issue of insufficient feature fine-tuning for downstream tasks. In contrast, re-embedding modules that can be end-to-end trained with MIL models provide supervised feature fine-tuning, which enables full exploitation of the knowledge beneficial to the final task. Hence, we can see from Table~\ref{tab:diff_trans} that any re-embedding structure achieves a considerable improvement on all tasks. Some tailored structures, such as our proposed R$^2$T, can have a significant performance advantage in most of the tasks. Moreover, Figure~\ref{fig:rt_bonus} shows that this improvement is not limited to the classical AB-MIL, but can widely benefit different MIL models. Table~\ref{tab:main_comp} demonstrates that the re-embedding approach is still effective on foundation model features. Therefore, compared to foundation model features, re-embedding features are a cheaper, more versatile alternative and effective booster.

\begin{table}[t]
\small
\centering
\begin{tabular}{lcccc}
\toprule
   Model         & C16$\uparrow$ & NSCLC$\uparrow$ & LUAD$\uparrow$ & TT$_{C16}\downarrow$\\ \midrule
AB-MIL+R50       & 94.52       & 95.28  & 58.78 &  3.1s\\
AB-MIL+PLIP       & 97.30       & 95.68  & 62.09 &  -\\
\textit{R50+Re-embedding} & &&\\
+TransMIL (global)     & 95.80       & 95.58  & 63.24 &  13.2s\\
+N-MSA (global)           & 96.20  & 95.51 & 63.99& 7.7s\\
+N-MSA (local)           & 96.47  & 95.97 & 65.41 & 29.8s\\
\rowcolor{dino}\textbf{+R$^2$T (local)}      & \textbf{97.32}       & \textbf{96.40} & \textbf{67.19}&  \textbf{6.5s} \\
\bottomrule
\end{tabular}
\caption{Comparison of different instance features under AB-MIL. 
We report the train time per epoch on C16 (TT$_{C16}$).
The pre-training time is not included for comparison.
}
\vspace{-0.3cm}
\label{tab:diff_trans}
\end{table}

\noindent\textbf{Different Re-embedding Discussion:}
The bottom part of Table~\ref{tab:diff_trans} presents the performance of AB-MIL under different feature settings. 
We employ four methods, including TranMIL, N-MSA, the local version of N-MSA, and our proposed R$^2$T to re-embed the features. 
From observations, all four re-embedded AB-MILs perform better than the original one. The performance improvement by TransMIL, N-MSA, and R$^2$T are 1.28\%, 1.68\%, and 2.80\%, respectively, on C16, while these numbers on LUAD are 4.46\%, 4.14\%, and 8.41\%, respectively. This phenomenon validates the importance of re-embedding in MIL-based computational pathology. Among the four employed re-embedding approaches, R$^2$T boosts the AB-MIL with the most considerable improvements while incurring the lowest computational cost. 
Specifically, R$^2$T achieves more performance gains (+0.85\% on C16 and +2.63\% on LUAD) while only requiring a 1/5 inference time compared with the second-best approaches. 



\begin{figure*}[t]
    \centering
    \includegraphics[width=14cm]{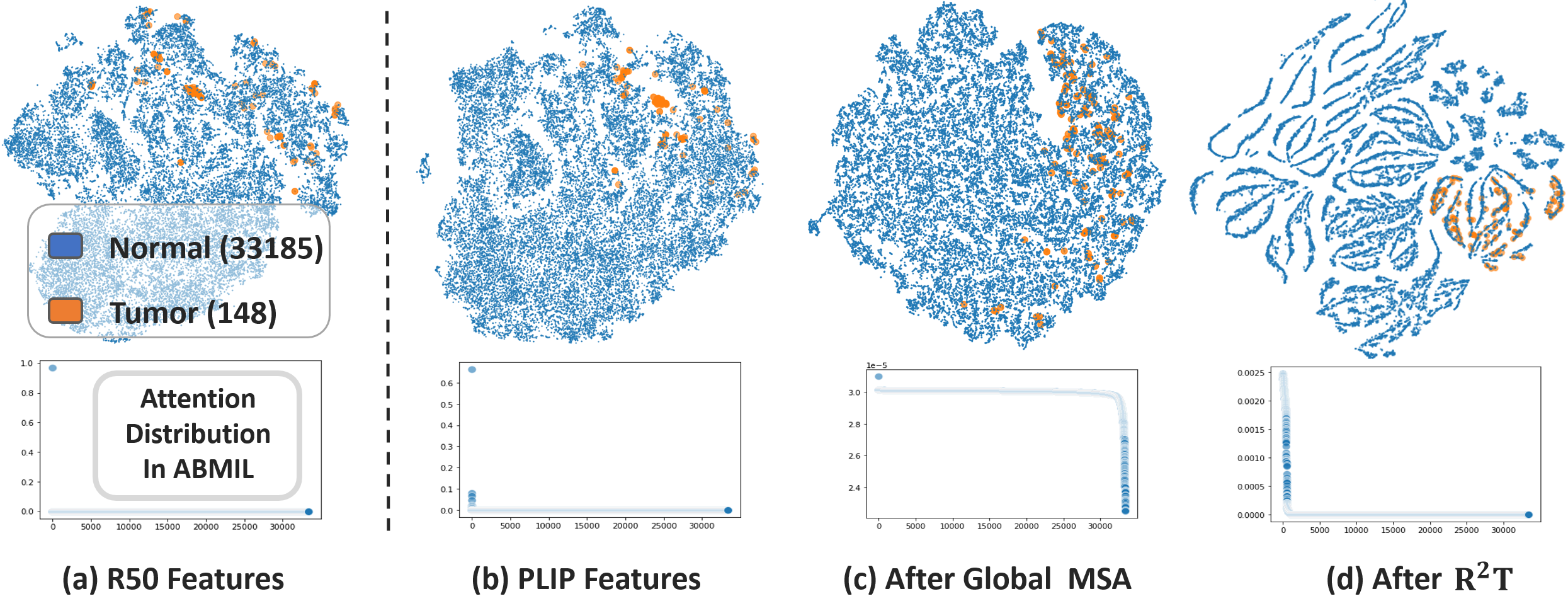}
    \caption{The tSNE~\cite{van2008tsne} visualization of instance features from the CAMELYON-16 dataset, comparing (a) features extracted by ResNet-50 pre-trained on ImageNet-1k, (b) features extracted by PLIP, (c) features after N-MSA re-embedding, and (d) features after R$^2$T re-embedding. In (a), we obtain instance-level labels from the tumor annotations and report the instance numbers of different labels.}
    \label{fig:intro_tsne}
\end{figure*}

\noindent\textbf{The Applicability of R$^2$T in MIL Frameworks:}
We incorporate the R$^2$T into different MIL frameworks as a re-embedding module for studying its applicability. 
The performance improvements achieved by this re-embedding module in different frameworks are shown in Figure~\ref{fig:rt_bonus}.
The results reveal that the R$^2$T is capable of improving all MIL frameworks on all tasks. 
Moreover, the improvement brought by R$^2$T surpasses the foundation model features on two tasks except for diagnosis, and reaches a similar level on C16.
This clearly verifies the good applicability of R$^2$T.

\subsubsection{Local is more Appropriate than Global}


To investigate the role of local self-attention in computational pathology, we replace the naive MSA in the partitioned regions with N-MSA~\cite{xiong2021nystromformer}. The results in Table~\ref{tab:diff_trans} demonstrate that feature re-embedding within local regions, called +N-MSA (local), outperforms the global method under the same N-MSA. This validates the superiority of local self-attention in mining fine-grained features over global ones. The cost of the performance improvement in local self-attention brings a new problem. The local ones suffer from a higher computational burden than the global ones (around 4$\times$ more training time). Our proposed R$^2$T can alleviate this problem since it employs a more naive MSA, which significantly reduces the computational cost. Another advantage of local self-attention is that it improves the diversity of the re-embedded features, as shown in Figure~\ref{fig:intro_tsne}~(c) and (d). Throughout, the local self-attention fashion is far more appropriate for re-embedding in computational pathology, and our proposed R$^2$T ensures both good performance and good efficiency in local self-attention.



\begin{table}[tb]
\small
\begin{center}
\begin{tabular}{lccc}
\toprule
   Model           & C16$\uparrow$ & NSCLC$\uparrow$ & LUAD$\uparrow$ \\ \midrule
w/o              & 96.82             & 96.01   & 65.45   \\
PEG$_{3\times3}$            &  96.86 (\textcolor{red}{+0.04})       &  96.11 (\textcolor{red}{+0.10})   & 65.61 (\textcolor{red}{+0.16})   \\
PEG$_{7\times7}$             & 95.47 (\textcolor{blue}{-1.35})       & 95.94 (\textcolor{blue}{-0.07})   & 65.14 (\textcolor{blue}{-0.31})   \\
PPEG            & 93.00 (\textcolor{blue}{-3.82})       & 96.03 (\textcolor{red}{+0.02})   & 65.28 (\textcolor{blue}{-0.17}) \\
\rowcolor{dino}\textbf{EPEG}           & \textbf{97.32 (\textcolor{red}{+0.50})}       & \textbf{96.40 (\textcolor{red}{+0.39})}  & \textbf{67.19 (\textcolor{red}{+1.74})}   \\
\bottomrule
\end{tabular}
\end{center}
\caption{Comparison results of different position encoding.
The PPEG~\cite{shao2021transmil} 
consists of a $3\times3$, $5\times5$, and $7\times7$ convolution block. 
}
\label{tab:diff_pe}
\end{table}

\subsubsection{Effects of EPEG}
We discuss the impact of various positional encoding methods that can handle variable input lengths in detail here.
Table~\ref{tab:diff_pe} shows that the conventional conditional positional encoding methods for the input, such as PEG~\cite{chu2021peg} and PPEG~\cite{shao2021transmil}, do not effectively improve the performance.
More parameters and more complex structures do not bring significant improvements, but rather the simplest PEG$_{3\times3}$ achieves slight improvements on both tasks. In contrast, the EPEG, which is embedded in MSA, can benefit from a more lightweight 1-D convolution, and encode the attention matrix more directly. This enables it to model the positional information more effectively in the re-embedding module. 
For instance, EPEG obtains 0.50\%, 0.39\%, and 1.74\% improvements on C16, NSCLC, and LUAD, respectively.

\subsubsection{Impact of Cross-Region MSA}
Although R-MSA can effectively mine the fine-grained features of local regions, the hard partitioning would restrict the range of re-embedding to each separate region. This impairs the discriminative power of the features, as they lack cross-region connections. The left figure in Table~\ref{tab:cr_msa} illustrates this phenomenon, where all features are divided into 64 clusters (corresponding to the number of regions). Even though one cluster can capture fine-grained features, the key clusters are scattered and not cohesive. This affects the expression of context-based semantic information, which is crucial for downstream tasks. The right figure shows the significant improvement of the feature distribution after adding the CR-MSA module, and the table results also prove its effectiveness. Moreover, such multi-region-based semantic information is more important for survival prediction than the other two tasks. This is because this task is based on cases, where each bag contains multiple slides, which requires a more comprehensive discrimination.

\begin{table}[t]
\centering
\includegraphics[width=7cm]{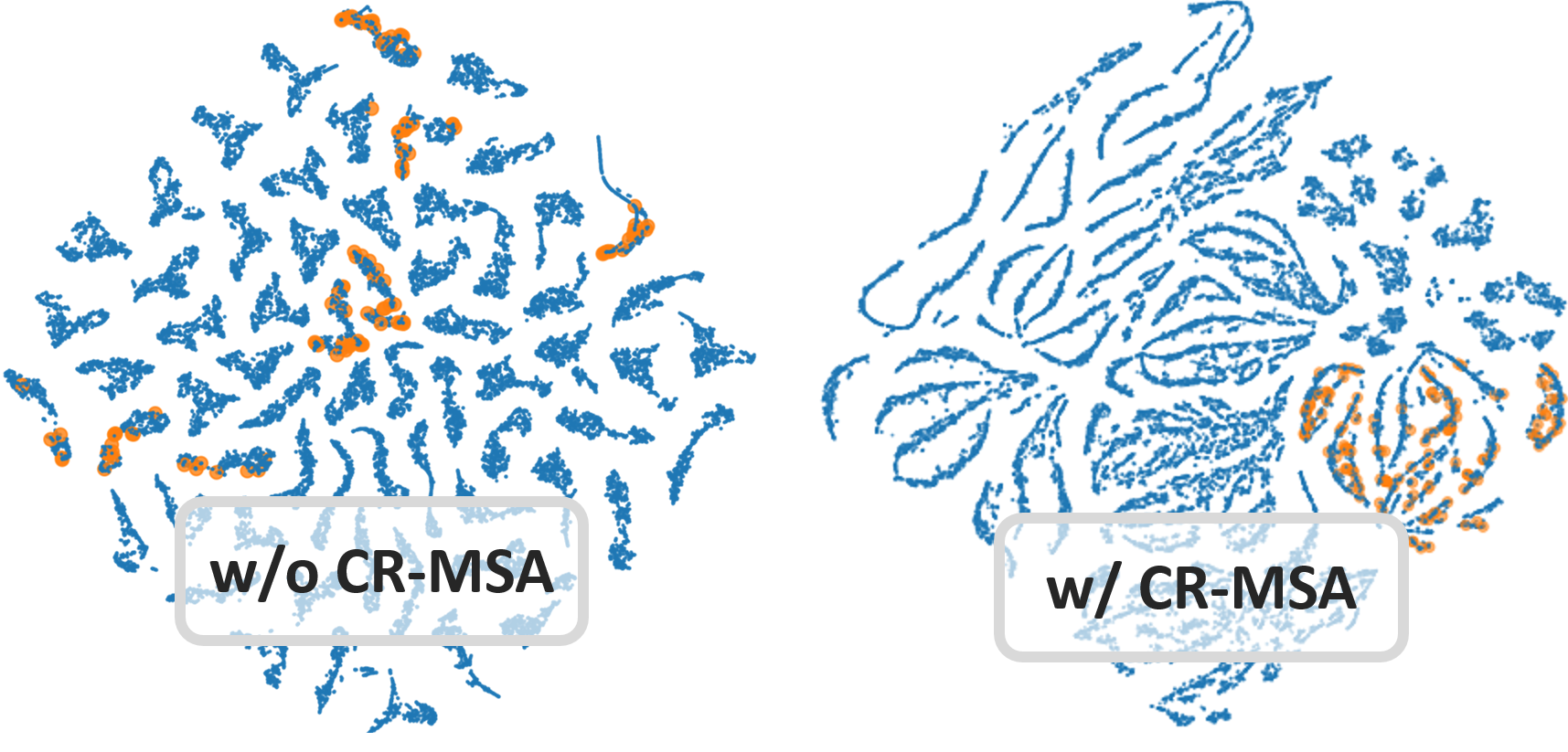}
\vspace{0.6cm}
\small
\begin{tabular}{lccc}
\toprule
Model   & C16$\uparrow$ & NSCLC$\uparrow$ & LUAD$\uparrow$ \\ \midrule
w/o   &  96.89   &   96.24   &   63.03   \\
\rowcolor{dino} \textbf{w/ CR-MSA}   &  \textbf{97.32 (\textcolor{red}{+0.43})}    &   \textbf{96.40 (\textcolor{red}{+0.16})}   &  \textbf{67.19 (\textcolor{red}{+4.16})}   \\ \bottomrule
\end{tabular}
\vspace{-0.5cm}
\caption{Quantitative and qualitative analysis of CR-MSA.}
\label{tab:cr_msa}
\vspace{-0.5cm}
\end{table}

\subsection{Visualization}
\label{sec:vis}
We visualize the features before and after re-embedding with different ways in Figure~\ref{fig:intro_tsne}. From observations, we can summarize that: 1) The offline feature extractor fails to learn the discriminative instance features, even with a foundation model trained on 200K slide-text pairs, especially when the instance distribution is extremely imbalanced, e.g., 1:224 positive-to-negative ratio;
2) Global self-attention enables the re-embedding of instance features, but its attention distribution is almost uniform. This indicates that its re-embedded features are homogeneous and lack diversity, which limits the performance of advanced MIL models.
3) The features re-embedded by R$^2$T not only enhance discriminability but also alleviate the issue of feature homogenization.
\section{Conclusion}
In this work, we have demonstrated the importance of instance feature re-embedding for computational pathology algorithms based on MIL, alleviating the issue of the under-learning of instance features in the conventional MIL paradigm. We have also shown that Transformer-based re-embedding modules can consistently boost the performance of various MIL methods regardless of their architectures.
However, the main result of this paper is the introduction of the Re-embedded Regional Transformer and two novel components: CR-MSA and EPEG.
We have evidence of the importance of the local Transformer in the age of the foundation model and its versatility as a re-embedding module.

\section{Acknowledgement}
Reported research is partly supported by the National Natural Science Foundation of China under Grant 62176030 and the Natural Science Foundation of Chongqing under Grant cstc2021jcyj-msxmX0568.
{
    \small
    \bibliographystyle{ieeenat_fullname}
    \bibliography{main}
}

\appendix
\SetAlFnt{\small}
\SetAlCapNameFnt{\small}

\begin{algorithm*}[t]
  \setstretch{0.8}
  \PyComment{x: input instance features} \\
  \PyComment{phi: learnable parameters $\Phi \in \mathbb{R}^{c \times k}$} \\
  \PyComment{msa: native MSA function} \\
  ~\\
  \PyComment{initialize} \\
  \PyCode{r, p, c = x.shape}\\
  \PyCode{logits = einsum("r p c, c k -> r p k", x, phi).transpose(1,2)}\\
  \PyComment{compute softmax weights} \\
  \PyCode{combine\underline{~}weights = logits.softmax(dim=-1)}\\
  \PyCode{dispatch\underline{~}weights = logits.softmax(dim=1)}\\
  \PyComment{compute minmax weights} \\
  \PyCode{logits\underline{~}min,\underline{~} = logits.min(dim=-1)}\\
  \PyCode{logits\underline{~}max,\underline{~} = logits.max(dim=-1)}\\
  \PyCode{dispatch\underline{~}weights\underline{~}mm = (logits - logits\underline{~}min) / (logits\underline{~}max - logits\underline{~}min + 1e-8)}\\
  \PyComment{get representative features of each region} \\
  \PyCode{x\underline{~}region = einsum("r p c, r k p -> r k p c", x,combine\underline{~}weights).sum(dim=-2)}\\
  \PyComment{perform native msa} \\
  \PyCode{z = msa(x\underline{~}region)}\\
  \PyComment{distribution of representative features} \\
  \PyCode{z = einsum("r k c, r k p -> r k p c", z, dispatch\underline{~}weights\underline{~}mm)}\\
  \PyComment{combination k of $\Phi$} \\
  \PyCode{z = einsum("r k p c, r k p -> r k p c", z, dispatch\underline{~}weights).sum(dim=1)}\\
  \Indm 

    \caption{PyTorch-style pseudocode for CR-MSA}
    \label{algo:train}
\end{algorithm*}

\section{Additional Method Detail}
\subsection{Attention Matrix}
Here, we further formulate the $e_{ij}^{l}$ of Equation~(6) in the manuscript as,
\begin{equation}
    e_{ij}^{l} = \frac{\left ( H^{l}_{i}W^{Q} \right ) \left ( H^{l}_{j}W^{K} \right )^{T}}{\sqrt{D}},
\end{equation}
where $H^{l}_{i}$ is the $i$-th instance features in $l$-the region $H^l$.
With EPEG, the R-MSA can be further represented as,
\begin{equation}
    \begin{aligned}
&\textrm{R-SA} = \sum_{j=1}^{M\times M}\alpha_{ij}^{l} \left ( H^{l}_{j}W^{V} \right ),\\
&\textrm{R-MSA} = \textrm{Concat}\left ( \textrm{R-SA}_1,\cdots,\textrm{R-SA}_{N_{head}} \right )W^O
\end{aligned}
\end{equation}
where the projections $W^{Q}$, $W^{K}$, $W^{V}$, and $W^{O}$ are parameter matrices sharing across the regions. The $N_{head}$ denotes the number of heads.

\subsection{Pseudocodes of CR-MSA}
Algorithm~\ref{algo:train} gives the details about CR-MSA.

\section{Dataset Description}
\noindent\textbf{CAMELYON-16~\cite{bejnordi2017diagnostic}} is a WSI dataset proposed for metastasis diagnosis in breast cancer.
The dataset contains a total of 400 WSIs, which are officially split into 270 for training and 130
~\footnote{Two slides in the test set are officially considered to be mislabeled, so they are not included in the experiment.}
for testing, and the testing sample ratio is 13/40$ \approx$1/3.
Following~\cite{Zhang_2022_BMVC,clam,chen2022hipt},
we adopt 3-times three-fold cross-validation on this dataset to ensure that each slide is used in training and testing, which can alleviate the impact of data split and random seed on the model evaluation. 
Each fold has approximately 133 slides. 
Although CAMELYON-16 provides pixel-level annotations of tumor regions, for weakly supervised learning, we only utilize slide-level annotations.

\noindent\textbf{TCGA NSCLC} includes two sub-type of cancers, Lung Adenocarcinoma (\textbf{LUAD}) and Lung Squamous Cell Carcinoma (\textbf{LUSC}). There are diagnostic slides, LUAD with 541 slides from 478 cases, and LUSC with 512 slides from 478 cases.
There are only slide-level labels available for this dataset. Compared to CAMELYON-16, tumor regions in tumor slides are significantly larger in this dataset.

\noindent\textbf{TCGA-BRCA} includes two sub-types of cancers, Invasive Ductal Carcinoma (IDC) and Invasive Lobular Carcinoma (ILC). There are 779 IDC slides and 198 ILC slides.
\noindent\textbf{TCGA-BLCA} contains 376 cases of Bladder Urothelial Carcinoma.

Following prior works~\cite{clam,shao2021transmil,zhang2022dtfd}, we crop each WSI into a series of $256 \times 256$ non-overlapping patches at 20X magnification. The background region, including holes, is discarded as in CLAM~\cite{clam}.

\section{Implementation Details}
Following~\cite{clam,shao2021transmil,zhang2022dtfd}, we use the ResNet-50 model~\cite{he2016deep} pretrained with ImageNet~\cite{deng2009imagenet} as the backbone network to extract an initial feature vector from each patch, which has a dimension of 1024. The last convolutional module of the ResNet-50 is removed, and a global average pooling is applied to the final feature maps to generate the initial feature vector. The initial feature vector is then reduced to a 512-dimensional feature vector by one fully-connected layer. 
As for PLIP~\cite{huang2023plip} features, we also use a fully-connected layer to map 512-dimensional features to 512 dimensions. The head number of R-MSA is 8.
An Adam optimizer~\cite{kingma2014adam} with learning rate of $2\times 10^{-4}$ and weight decay of $1\times 10^{-5}$ is used for the model training. The Cosine strategy is adopted to adjust the learning rate.
All the models are trained for 200 epochs with an early-stopping strategy.
The patience of CAMELYON-16 and TCGA are 30 and 20, respectively.
We do not use any trick to improve the model performance, such as gradient cropping or gradient accumulation.
The batch size is set to 1. All the experiments are conducted with NVIDIA GPUs.
Section~\ref{sec:code} gives all codes and weights of the pre-trained PLIP model.

\begin{figure}[t]
    \begin{center}
    \includegraphics[width=0.9\linewidth]{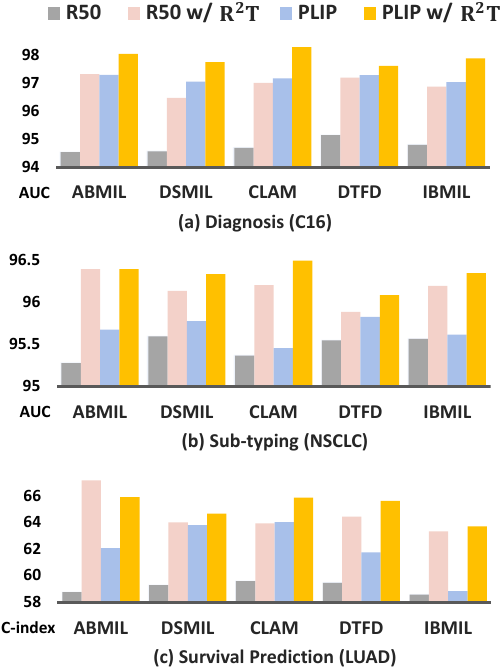}
    \end{center}
    \caption{Performance improvement by adding R$^2$T on different offline features.}
    \label{fig:rt}
\end{figure}

\section{Additional Quantitative Experiments}

\subsection{More on Foundation Model Features}


In this section, we evaluate the improvement of R$^2$T on foundation model features with more MIL models. Figure~\ref{fig:rt} shows the advantage of R$^2$T by online fine-tuning, which can further enhance the discriminability of foundation model features on multiple tasks and models. Surprisingly, we find that this improvement does not decrease with more advanced MIL models, such as R$^2$T+CLAM often outperforms R$^2$T+ABMIL.

\subsection{More on EPEG}

\begin{figure*}[t]
    \centering
    \includegraphics[width=15cm]{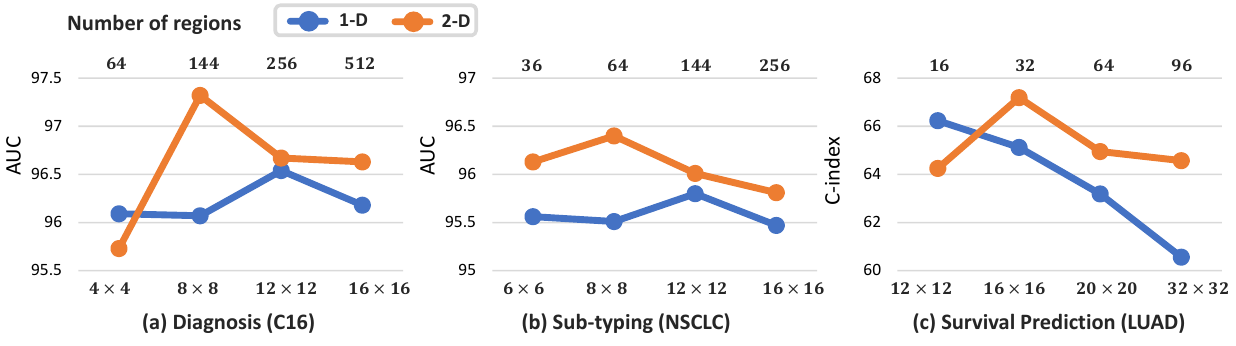}
    \caption{The performances under different region partition strategies on two datasets.}
    \label{fig:reginal_division}
\end{figure*}

\noindent\textbf{Different Convolution Kernel.}
Here, we discuss the impact of different convolution kernels on EPEG. The $k$ is the optimal kernel size, and the upper part of Figure~\ref{fig:hyper_para} discusses more on different datasets. First, we find that most 1-D convolutional kernels enhance the re-embedding ability of local Transformers. Second, larger convolution kernels typically perform worse. We attribute this to the excessive parameters that tend to overfit on a limited number of slides.  


\begin{table}[hb]
\small
\begin{center}
\begin{tabular}{lcccc}
\toprule
   Type  & Kernel        & C16 & NSCLC & LUAD \\ \midrule
w/o       &  none    & 96.82             & 96.01  & 65.45\\
2-D       &   3$\times$3   &   96.73     & 95.93  & 66.70\\
2-D       &   7$\times$7   &    96.60    & 95.71  & 64.59\\
\rowcolor{dino}\textbf{1-D}       &   \textbf{k$\times$1}  & \textbf{97.32}    & \textbf{96.40}  & \textbf{67.19}\\
\bottomrule
\end{tabular}
\end{center}
\label{tab:diff_pe_k}
\end{table}

\begin{figure}[t]
    \begin{center}
    \includegraphics[width=\linewidth]{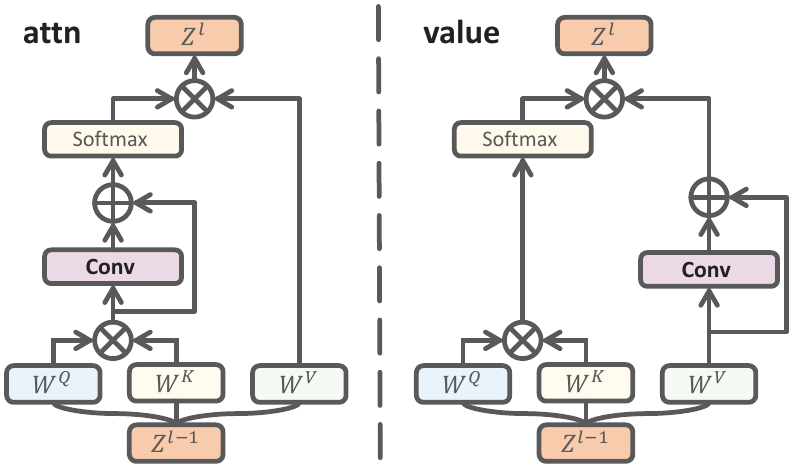}
    \end{center}
    \caption{Illustration of variants of EPEG. The left one is the default.}
    \label{fig:epeg_2}
\end{figure}

\noindent\textbf{Different Embedded Position.}
Here, we discuss another variant of EPEG. We place the convolution module after the “value” matrix instead of the default “attn” matrix. Figure~\ref{fig:epeg_2} shows the specific structures of the two variants. The results in Table~\ref{tab:epeg_2} demonstrate the feasibility of the “value” variant, but its performance is significantly lower than the original version, especially on the more challenging C16 dataset. We note this is because the original version can incorporate positional information into the core attention matrix to model positional information more directly.

\begin{table}[htb]
\begin{center}
\begin{tabular}{lcccc}
\toprule
   Type  & Kernel        & C16 & NSCLC & LUAD\\ \midrule
w/o       &  none    & 96.82             & 96.01  & 65.45\\
value      &   3$\times$3   &  96.78      &  96.07 & 64.47\\
value      &   k$\times$1   &  96.90      & 96.10  & 65.76\\
\rowcolor{dino}\textbf{attn}       &   \textbf{k$\times$1}   & \textbf{97.32}       & \textbf{96.40}  & \textbf{67.19}\\
\bottomrule
\end{tabular}
\end{center}
\caption{Comparison results of variants of EPEG. }
\label{tab:epeg_2}
\end{table}

\subsection{Region Partition Strategy}
Here, we systematically investigate the impacts of different region partition strategies in our method. Figure~\ref{fig:reginal_division} reports the computational pathology performances of our method with different region partition settings. From observations, we find that the 2-D region partition fashion is superior to the 1-D ones because it preserves more original image structure information. Another phenomenon is that employing a too-small or too-large region will degenerate the performances. We attribute this to the fact that the small region sharply reduces the module receptive field, while the large region damages the diversity of the re-embedded features. Therefore, a moderate region is optimal for R-MSA. 
Furthermore, even with a larger number of region partitions (e.g., 512 or 32$\times$32), the proposed model still maintains a high level of performance. This reflects that our model can achieve a good trade-off between performance and efficiency (more regions resulting in lower spatial and temporal costs). These findings demonstrate the good scalability of our model to the datasets that have longer input sequences.

\begin{table*}[t]
\small
\begin{center}
\begin{tabular}{lcccccccc}
\toprule
       & PretrainingData$\downarrow$  & Para.$_{+\textrm{offline}}\downarrow$ &TrainTime$\downarrow$ & Memo.$\downarrow$ & FPS$\uparrow$ &C16$\uparrow$ & NSCLC$\uparrow$& LUAD$\uparrow$  \\ \midrule
ABMIL        & ImageNet-1K  &0.65M$_{+26M}$ & 3.1s &2.3G & 1250 & 94.54  & 95.28 & 58.78 \\ 
ABMIL+PLIP [11]   & \textbf{\textcolor{red}{OpenPath-200K}} &0.65M$_{\textbf{\textcolor{red}{+151M}}}$ & 1.7s &2.2G & 2273 & \underline{97.30}  & 95.68 & 62.09 \\
DTFD [41]        & ImageNet-1K  &0.79M$_{+26M}$ & 5.1s &2.1G & 325 & 95.15  & 95.55 & 59.48 \\
TransMIL [22]     & ImageNet-1K  & \textbf{\textcolor{red}{2.67M}}$_{+26M}$ & \textbf{\textcolor{red}{13.2s}} & \textbf{\textcolor{red}{10.6G}} & \textbf{\textcolor{red}{76}} & 93.51  & 94.97 & 64.11 \\
\midrule
\textit{Re-embedding}&&&&&&&\\
ABMIL+N-MSA [36]    & ImageNet-1K  &1.64M$_{+26M}$ & 7.7s &7.2G & 158 & 96.20  & 95.51 & 63.99 \\
\rowcolor{dino}R$^2$T-MIL(w/o CR-MSA) & ImageNet-1K &1.64M$_{+26M}$ & 6.1s &10.0G & 272 & 96.89  & \underline{96.24} & 63.03 \\
\rowcolor{dino}R$^2$T-MIL& ImageNet-1K &2.70M$_{+26M}$ & 6.5s &10.1G & 236 & \textbf{97.32}  & \textbf{96.40} & \textbf{67.19} \\ 
R$^2$T-MIL [w/ FFN]$_{x2}$ & ImageNet-1K &6.90M$_{+26M}$ & 9.9s &12.0G & 114 & 96.23  & 95.58 & \underline{64.89} \\ 
R$^2$T-MIL [w/ FFN]$_{x3}$ & ImageNet-1K &10.05M$_{+26M}$ & 14.4s &16.3G & 71 & 96.57  & 95.70 & 63.07 \\ 
 
\bottomrule

\end{tabular}
\end{center}
\caption{More about the efficiency analysis of R$^2$T. FFN denotes the feed-forward network. We use subscripts to indicate the number of layers. It is worth noting that the feature dimensions of the remaining terms except PLIP features are 1024, while PLIP is 512, which explains its rise in efficiency compared to ABMIL. Other than that, the PLIP feature does not have any computational cost impact on the original method.}
\label{tab:ly}
\end{table*}

\subsection{More Parameters are not Always Better}

\begin{figure}[t]
    \begin{center}
    \includegraphics[width=7cm]{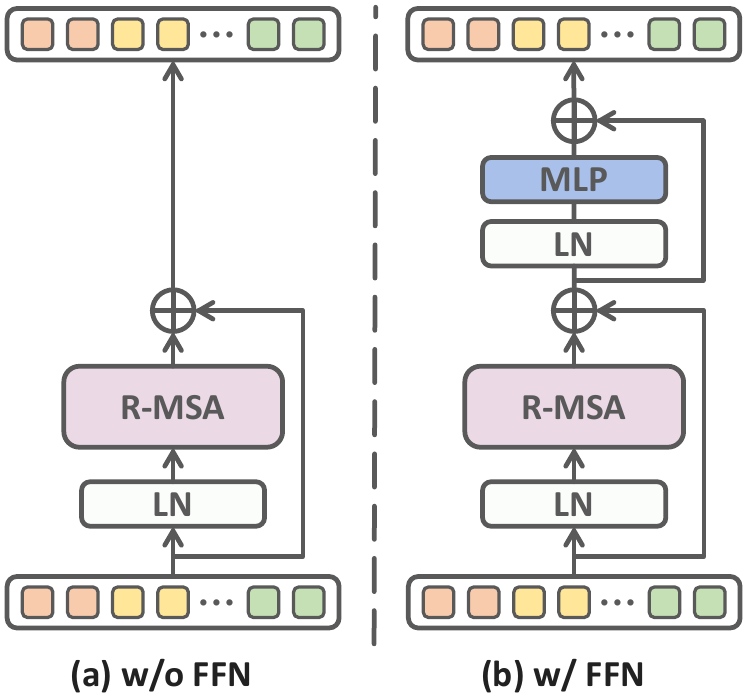}
    \end{center}
    \caption{Illustration of different blocks of R$^2$T. The (a) is the default.}
    \label{fig:ffn}
\end{figure}

Generally, a Transformer is a multi-layer structure that stacks several blocks with the same structure~\cite{liu2021swin,vaswani2017attention,lan2019albert,dosovitskiy2020image}. 
However, due to the task specificity, R$^2$T only contains a few blocks. Here, we systematically investigate the impact of different numbers of layers and different blocks on the performance and computational cost of R$^2$T. 
First, Figure~\ref{fig:ffn} shows two different blocks: (a) is used by default in R$^2$T; (b) introduces a feed-forward network (FFN), which plays an indispensable role in Transformer for NLP or natural image computer vision tasks. From Table~\ref{tab:ly}, we can find that FFN introduces a large number of parameters and computation, but the more expensive computation cost does not bring performance improvement.

Moreover, 
we can summarize that: 1) Transformer-based methods, represented by TransMIL, bring better long-sequence modeling ability, but also introduce several times more parameters. R$^2$T-MIL, as one of them, has equal or less parameter size compared to TransMIL and better performance. 2) As Transformer-based methods, Re-embedding paradigm and R$^2$T have Higher parameter efficiency. ``+N-MSA" has the same model structure as TransMIL (N-MSA$\times$2), but thanks to the Re-embedding, it can achieve higher performance with lower cost. Moreover, R$^2$T (w/o CR-MSA) leverages a more excellent design to further improve performance and training time. This indicates that R$^2$T-MIL has a good parameter compression space, and can achieve significant improvement with 2$\times$ parameters compared to DTFD, reaching the level of the foundation model. 
3) In computational pathology, limited by the number of slides, the models face the problem of over-fitting\cite{zhang2022dtfd}, and higher parameter size does not imply better performance.
Not only does TransMIL perform poorly on some tasks, we also 
add FFN and increase the number of layers on the R$^2$T-MIL to 
increase the parameters of R$^2$T-MIL (+7.35M), but the performance drops significantly (-4.12\% on LUAD).


\subsection{More on Local Transformer}
Table~\ref{tab:local_trans} further explores the impact of local Transformer on computational pathology performance. We set a threshold and perform global MSA computation instead of regional MSA for bags with instance numbers less than that threshold. 
First, we can find that more use of global MSA leads to worse performance on both datasets. 
The characteristic of small tumor areas on the C16 dataset exacerbates the performance degradation caused by global MSA. 
In addition, this strategy introduces extra hyperparameters, reducing the generalization ability of the model. 
Overall, our experiments prove that local Transformers can better adapt to the inherent characteristics of WSI such as huge size and small tumor areas than traditional global Transformers.
\begin{table}[htb]
\begin{center}
\begin{tabular}{lccc}
\toprule
   case           & C16 & NSCLC  & LUAD \\ \midrule
\rowcolor{dino}\textbf{$\leq$0}              & \textbf{97.32}             & \textbf{96.40}  &  \textbf{67.19} \\
$\leq$500             &  96.99 (\textcolor{blue}{-0.33})       &  96.23 (\textcolor{blue}{-0.17})  &   62.35 (\textcolor{blue}{-4.84})  \\
$\leq$1000             & 97.21 (\textcolor{blue}{-0.11})       & 95.98 (\textcolor{blue}{-0.42})  &  62.15 (\textcolor{blue}{-5.04})  \\
\bottomrule
\end{tabular}
\end{center}
\caption{Comparison results between global MSA and local MSA. We perform global MSA computation instead of regional MSA for bags with instance numbers less than the threshold.}
\label{tab:local_trans}
\end{table}

\subsection{Discussion of Hyper-parameter in CR-MSA}
\begin{figure*}[t]
    \begin{center}
    \includegraphics[width=16.5cm]{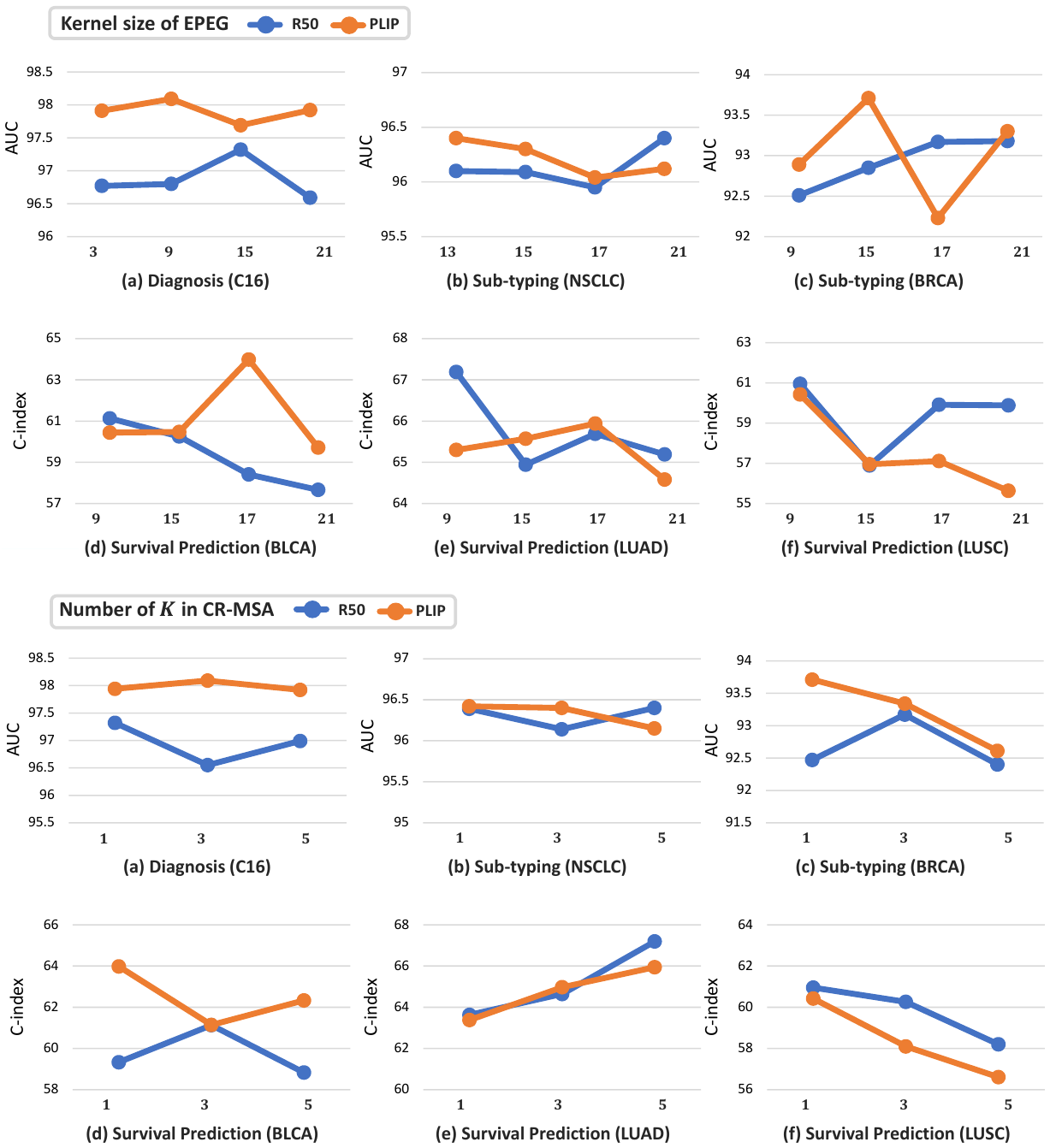}
    \end{center}
    \caption{Discussion of important hyper-parameters.}
    \label{fig:hyper_para}
\end{figure*}

The bottom part of Figure~\ref{fig:hyper_para} shows the results. We can find that R$^2$T is not sensitive to this parameter, and different values can achieve high-level performance. In addition, different offline features show similar consistency. This reflects the generality of the preset optimal parameter in different scenarios.

\section{Additional Visualization}
\begin{figure*}[t]
    \begin{center}
    \includegraphics[width=14cm]{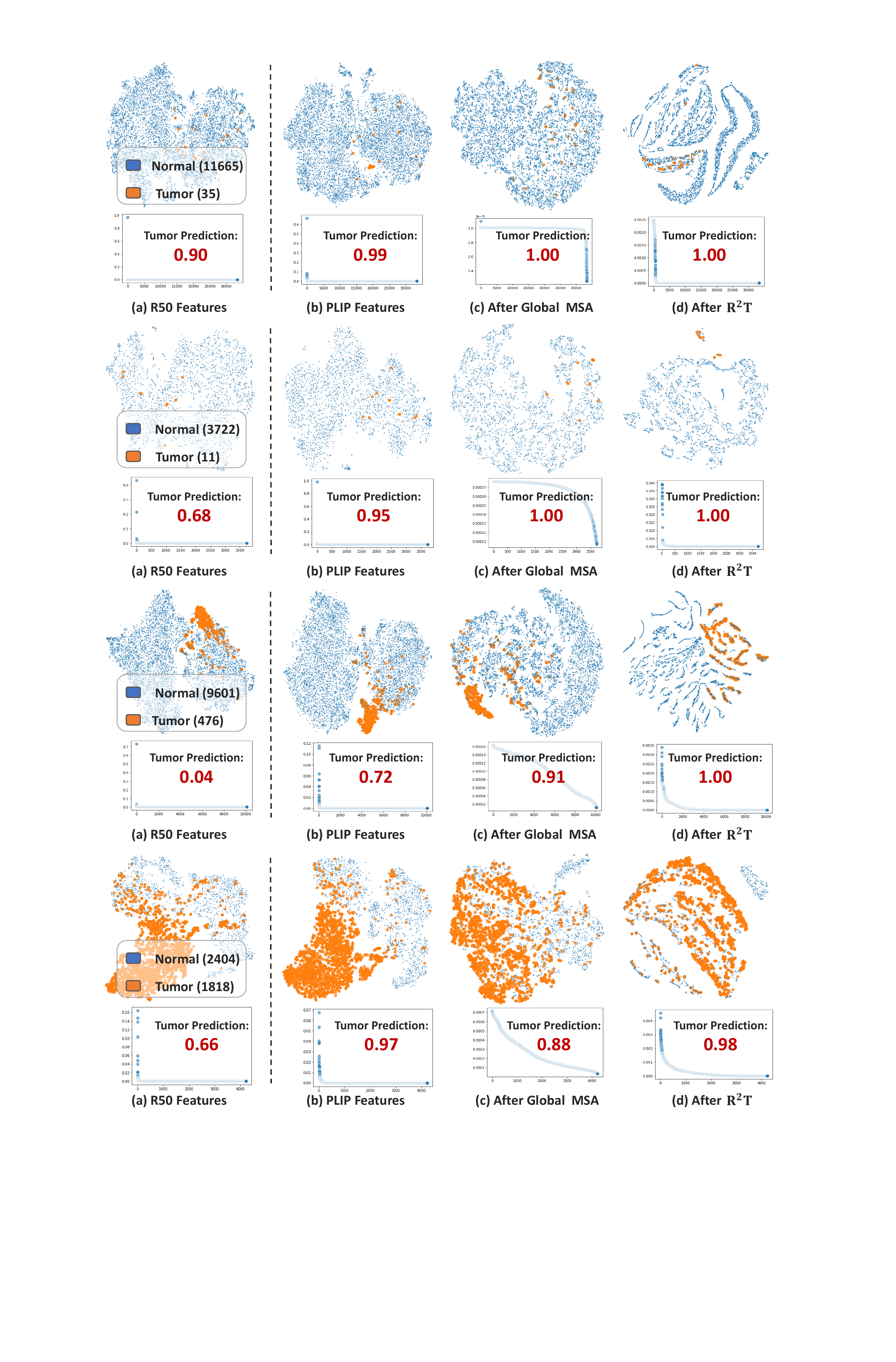}
    \end{center}
    \caption{More comparison of t-SNE visualization of instance features. Best viewed in scale.}
    \label{fig:big_vis}
\end{figure*}

Figure~\ref{fig:big_vis} presents more comprehensive feature visualizations, including cases where the original features have high and extremely low discriminativeness. We use the attention score after softmax normalization to label instances for demonstrating the updated features. Moreover, when the tumor prediction confidence is too low, we assume that the attention score cannot directly indicate the instance tumor probability. In this case, we still use original instance labels to colorize the visualization. 

From Figure~\ref{fig:big_vis}, we can draw the following conclusions: 
1) Although the final MIL model can correctly classify the slide with high discriminator original features, the feature visualization after linear projection (row 1) is still unsatisfactory (high coupling and unclear cohesion). 
2) Features with low discriminativeness (rows 2 and 3) impair the judgment of the MIL model, and the re-embedding module can effectively enhance feature discriminativeness. 
3) In features re-embedded by global MSA, the distribution area of tumor instances is linearly correlated with the final tumor prediction score. The larger the distribution area of tumor instances, usually higher its tumor prediction score. 
However, too many instance numbers and an extremely low tumor instance ratio make it difficult for the module to re-embed all instances as tumor instances, which ultimately affects the performance of the MIL model. We attribute this to a lack of diversity in features re-embedded by global MSA. 
4) In contrast, regional MSA addresses this problem well. 
Because features among different regions are distinct from each other, even if the proportion of re-embedded tumor instances is still low, their discriminativeness is very high (high cohesion and low coupling), which is more favorable for the classification of the final MIL model.



\section{Limitation}
\label{sec:limitaion}
Although the Transformer-based re-embedding module can effectively improve the discriminativeness of instance features and facilitate classification, we find that the re-embedded features lose their original label information due to the self-attention update of the original features. For example, an original non-tumor patch may be re-embedded as a tumor patch to benefit slide classification. This characteristic of the re-embedding module prevents it from accurately performing weakly supervised localization or segmentation of tumor areas through the final aggregation module. However, the outstanding weakly supervised localization and segmentation capability is one of the features of attention-based MIL models. Therefore, how to use the re-embedding module to improve detection or segmentation performance is our future work.

\section{Code and Data Availability}
\label{sec:code}
The source code of our project will be uploaded at~\href{https://github.com/DearCaat/RRT-MIL}{https://github.com/DearCaat/RRT-MIL}.

CAMELYON-16 dataset can be found at~\href{https://camelyon16.grand-challenge.org}{https://camelyon16.grand-challenge.org}.

All TCGA datasets can be found at~\href{https://portal.gdc.cancer.gov}{https://portal.gdc.cancer.gov}.

The script of slide pre-processing and patching can be found at~\href{https://github.com/mahmoodlab/CLAM}{https://github.com/mahmoodlab/CLAM}.

The code and weights of PLIP can be found at~\href{https://github.com/PathologyFoundation/plip}{https://github.com/PathologyFoundation/plip}.

\end{document}